\documentclass[conference]{IEEEtran}
\IEEEoverridecommandlockouts
\usepackage{cite}
\usepackage{hyperref}
\usepackage{amsmath,amssymb,amsfonts}
\usepackage{algorithmic}
\usepackage{graphicx}
\usepackage{textcomp}
\usepackage{xcolor}
\usepackage{listings}
\usepackage{booktabs}
\usepackage{stfloats}

\usepackage{multirow}
\def\BibTeX{{\rm B\kern-.05em{\sc i\kern-.025em b}\kern-.08em
    T\kern-.1667em\lower.7ex\hbox{E}\kern-.125emX}}
\begin{document}

\title{

From Language to Navigation Goals: A Vision-Language Approach for Semantic Navigation of Mobile Robots Using RGB-D Perception\\

\thanks{
This work has been supported by MCIN/AEI/10.13039/501100011033, under project AI-FUSE-ROBOT (SAIA202500X163851SV0) and by CIN/AEI/10.13039/501100011033 and by ERDF "A way of making Europe" under project COBUILD: Collaborative team of robots and humans for advanced digital building (PID2024-161069OB-I00) and Juan de la Cierva grant JDC2024-054576-I from MICIU/AEI/10.13039/501100011033.
}
}

\author{
\IEEEauthorblockN{Jose Martinez Fajardo}
\IEEEauthorblockA{
\textit{Univ. Pablo de Olavide}\\
jmarfaj1@alu.upo.es}
\and
\IEEEauthorblockN{Pablo Pueyo}
\IEEEauthorblockA{
\textit{Univ. Pablo de Olavide}\\
ppueram@upo.es}
\and
\IEEEauthorblockN{Fernando Caballero}
\IEEEauthorblockA{
\textit{Univ. Pablo de Olavide}\\
fcaballero@upo.es}
\and
\IEEEauthorblockN{Luis Merino}
\IEEEauthorblockA{
\textit{Univ. Pablo de Olavide}\\
lmercab@upo.es}
}
\maketitle

\begin{abstract}


Natural language interaction provides an intuitive
way for non-expert users to communicate with robotic platforms.
However, transforming user requests into executable navigation
actions remains a challenging task, requiring the integration
of language understanding, environment perception, and autonomous navigation. This work presents a language-driven
navigation framework that enables mobile robots to interpret
user requests in natural language to move the robot to a
destination, and autonomously navigate towards it.
Given a natural language request referring to
a target in the environment (e.g. “go to the mail box”), the system
identifies the referenced object, estimates its position using RGBD data, and generates a navigation goal, which is then executed
through the ROS 2 Nav2 navigation stack. The ROS 2-based
implementation facilitates portability across different robotic
platforms, requiring only the configuration of the corresponding
topics and services.
The system is evaluated in both simulation and real-world
scenarios using a TurtleBot3 Waffle and a Unitree Go2 robot
with a Realsense camera. Experimental results show that the
framework successfully interprets both direct commands and
contextual requests, generates meaningful natural-language feedback, and navigates towards the desired target. These results
demonstrate the feasibility of combining semantic perception
and autonomous navigation to provide an intuitive human-robot
interaction paradigm. \footnote{The source code for this project is available at \url{https://github.com/tu-usuario/tu-repositorio}}.

\end{abstract}


\section{Introduction}
The increasing democratization of mobile robots beyond
research and industrial environments is creating a need for
intuitive human–robot interaction mechanisms. As robotic
platforms become accessible to non-expert users, natural language provides a user-friendly communication
modality, allowing users to specify tasks without requiring
knowledge of robot programming, coordinates, or navigation
systems.

Traditional navigation systems typically rely on explicit
goal definitions, including metric coordinates, manually selected waypoints, or predefined semantic maps. Although
these approaches provide reliable and accurate navigation,
they are limited to expert users. A fundamental challenge
in this context is language-driven navigation.
Recent advances in Vision-Language Models (VLMs) provide a promising solution to this problem. By processing
visual and textual information, VLMs can associate linguistic
concepts with visual elements in a scene, enabling robots to
identify objects and regions referenced in natural language
commands. Combined with RGB-D perception, these models
can be used not only to identify the target object referenced
by the user but also to estimate its position in the physical
environment.
In this work, we present a language-driven navigation
framework that combines VLM-based semantic perception,
RGB-D sensing, and autonomous navigation. Given a natural
language request, the system identifies the referenced target,
estimates its three-dimensional position through geometric
projection, and generates a navigation goal that can be executed by a standard robotic navigation stack. The framework is
implemented in ROS 2 and leverages the Navigation2 (Nav2)
stack for autonomous navigation, enabling deployment on
different robotic platforms equipped with an RGB-D camera.
The main contributions of this work are summarized as follows:
\begin{itemize}
    \item A language-driven navigation framework that transforms natural language requests into executable navigation goals through the integration of semantic perception and autonomous navigation.
    \item A method for grounding VLM detections in metric space using RGB-D data, enabling the generation of physically meaningful navigation objectives.
    \item The implementation and validation of a complete ROS~2-based pipeline interpreting commands in simulation, complemented by a real-world proof of concept for semantic perception positioning.
\end{itemize}

To promote reproducibility and facilitate further research in
language-driven robotics, the source code associated with this
work will be released as open source upon acceptance.
The remainder of this paper is organized as follows. Section II reviews related work, Section III presents the proposed
framework, Section IV describes the experimental evaluation,
and Section V concludes the paper.

\begin{figure*}[!th]
\centering
\includegraphics[width=0.8\textwidth]{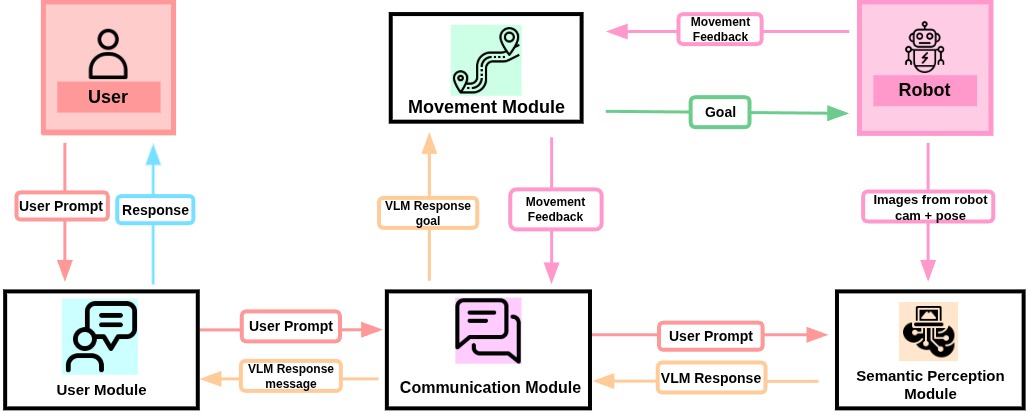}
\caption{\textbf{System diagram: }The user provides natural language prompts to the system through the User module. The robot sends RGB-D images to the Communication module and, in addition, sends navigation feedback to and receives a navigation goal from the Movement module. The Communication module forwards the robot images and the user prompt to the VLM module, which returns both a navigation goal and a natural language response. The Movement module acts as a mediator that receives the goal from the Communication module and publishes it to the robot's movement action. Finally, the VLM response is sent to the user through the User module, which receives it from the Communication module.}
\label{fig:diagram}
\end{figure*}
\footnotetext{Link to the repository on github: https://github.com/jmarfaj10/Vision-Language-Approach-for-Semantic-Navigation}
\section{Related Work}
\label{sec:related}

Recent advances in Vision-Language Models (VLMs) have significantly improved the ability to associate natural language descriptions with visual observations. Models such as CLIP \cite{radford2021clip} and BLIP-2 \cite{li2023blip2} learn joint visual-textual representations that support semantic understanding, visual grounding, and multimodal reasoning. These capabilities have motivated their adoption in robotics, where language and visual perception must be combined to understand human intentions.

Several works have explored the integration of foundation models into robotic systems. SayCan \cite{ahn2022saycan} combines large language models with robot affordances to generate executable action plans, while RT-2 \cite{zitkovich2023rt} and PaLM-E \cite{driess2023palme} demonstrate the potential of unified vision-language-control architectures for embodied tasks.
Similarly, $\pi_{0.5}$ \cite{black2025pi} extends vision-language-action models with improved open-world generalization, enabling robots to perform complex tasks in previously unseen environments.

Natural language interaction has been extensively studied in
Vision-and-Language Navigation (VLN), where agents must
follow navigation instructions using visual observations \cite{anderson2018vision}. 

The introduction of the Room-to-Room benchmark established
a standard framework for evaluating navigation from natural
language descriptions and inspired a large body of subsequent
research. More recently, surveys such as \cite{zhou2024vision} have identified
language grounding, generalization, and robustness as the primary challenges in VLN systems. Although these approaches
have demonstrated promising results in benchmark environments, many of them rely on task-specific datasets, endto-end learned policies, or simulation-oriented settings that
are difficult to integrate into conventional robotic navigation
frameworks.

A key challenge in language-driven navigation is converting semantic references into physically meaningful navigation goals. Early works explored the mapping of language instructions into actions and visual goals \cite{misra2017mapping}, while more recent approaches have leveraged vision-language representations to construct semantic maps and support object-centric navigation. For example, VLMaps \cite{huang2023vlmaps} uses VLM features to enable robots to localize semantically described objects within an environment. Complementing semantic understanding, RGB-D sensing remains one of the most widely adopted perception modalities in mobile robotics due to its ability to provide geometric information for localization and navigation \cite{zhang2023rgbd}. Combined with classical geometric projection methods \cite{hartley2003multiple}, RGB-D data enable visual detections to be transformed into metric coordinates that can be processed by navigation systems such as Nav2 \cite{macenski2020marathon}.

In contrast to existing VLN and embodied AI approaches, this work focuses on a lightweight and portable ROS~2 framework that directly transforms natural language requests into executable navigation goals. The proposed approach combines VLM-based semantic understanding with RGB-D geometric localization to estimate target positions in metric space and generate goals compatible with standard navigation systems. This design enables deployment across different robotic platforms while preserving compatibility with existing ROS~2 and Nav2 infrastructures.

\section{System Architecture}
\label{sec:system}

Figure~\ref{fig:diagram} presents the architecture of the proposed language-driven navigation framework, organized around two main entities: the \textit{user} and the \textit{robot}. The user provides navigation requests through natural language, while the robot perceives the environment and executes the requested actions. To bridge both entities, the system is composed of four main modules: the User Module, the Communication Module, the Semantic Perception Module, and the Movement Module. Implemented in ROS~2, these modules cooperate to transform natural language requests into navigation actions through semantic perception, goal generation, and autonomous navigation, while also providing natural-language feedback to the user. The ROS~2-based architecture facilitates deployment across different robotic platforms equipped with RGB-D sensors.

\label{sec:infrastructure}

\subsection{User Module}
The User Module provides the interface between the user
and the robotic system, enabling navigation goals to be
specified through natural language rather than coordinates or
predefined waypoints. User commands are forwarded to the
Communication Module for processing.

Typical requests include commands such as \textit{``go to the refrigerator''}, \textit{``move to the door''}, or \textit{``navigate to the chair''}. After processing the instruction, the system generates a natural-language response confirming the identified destination. This feedback allows the user to verify that the request has been correctly interpreted before navigation begins.

By abstracting low-level navigation details, the User Module provides an intuitive interaction mechanism that facilitates the use of mobile robots by non-expert users.

\subsection{Communication Module}

The Communication Module acts as the central coordinator
of the architecture and implements the execution logic of the
system. All interactions between the different modules are
managed through this component, which is responsible for
controlling the information flow and determining when data
must be acquired, processed, and transmitted throughout the
navigation pipeline.

This module also provides an abstraction layer between the software architecture and the underlying robotic platform. This capability is enabled by ROS~2, which serves as the communication backbone of the system through topics, services, and actions. By translating robot-specific interfaces into system-level interfaces, the Communication Module allows the remaining modules to operate independently of the particular hardware configuration. As a result, the proposed framework can be deployed on different robotic platforms equipped with RGB-D sensors by adapting only the corresponding ROS~2 topic and action interfaces.

When a navigation command is received from the User Module, the Communication Module acquires the RGB and depth information provided by the robot sensors and forwards the corresponding request to the VLM Module. Once the target has been identified and localized, the resulting navigation objective is delivered to the Movement Module for execution. In addition, the module collects feedback from both the VLM and navigation processes.

\subsection{Semantic Perception Module}
\label{sec:semantic}
The Semantic Perception Module connects natural language understanding and robot perception. Its primary objective is to establish communication with the server hosting the vision-language model (VLM); in this case, the hosted model is QWEN3, though any other model can be used.

Upon receiving a request from the Communication Module, the Semantic Perception Module sends both the user's natural language instruction and the RGB image captured by the robot to the remote VLM server. By jointly analyzing the visual and textual information, the model identifies the object or scene element associated with the user's request and returns its location in the image as a bounding box. The detected bounding box is defined by the image points $\mathbf{p}_{tl}\in\mathbb{R}^{2}$ and $\mathbf{p}_{br}\in\mathbb{R}^{2}$, corresponding to its top-left and bottom-right corners, respectively. Additionally, the VLM generates a natural language response, such as \textit{``Okay, I will go to the chair''}, which is forwarded to the user to confirm that the instruction has been correctly interpreted.

\begin{figure}[t]
    \centering
    \includegraphics[width=0.48\linewidth, height=2.5cm]{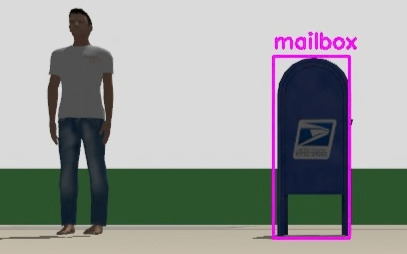}
    \hfill
    \includegraphics[width=0.48\linewidth, height=2.5cm]{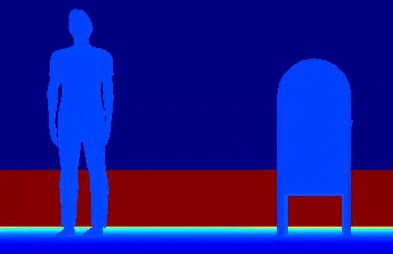}
    \caption{Semantic target localization process for the command \textit{``go to the mailbox''}. Left: RGB image showing the bounding box returned by the VLM around the detected mailbox, defined by the points $\mathbf{p}_{tl}$ and $\mathbf{p}_{br}$. The center point $\mathbf{p}=(p_x,p_y)$ is selected as the representative image point associated with the detected target. Right: Corresponding depth image. A local neighborhood centered at $\mathbf{p}$ is used to estimate the target depth by selecting the minimum valid depth value within the region.}
    \label{fig:semantic_localization}
\end{figure}

Figure~\ref{fig:semantic_localization} illustrates an example of the semantic perception process for the command \textit{``go to the mailbox''}. The VLM identifies the mailbox in the RGB image and returns the corresponding bounding box. The resulting detection is then combined with the depth image provided by the RGB-D sensor to estimate the three-dimensional position of the target.

Since the subsequent geometric processing requires a single image point, the center of the bounding box is selected as a representative target location:

\begin{equation}
\mathbf{p}=(p_x,p_y)=\frac{\mathbf{p}_{tl}+\mathbf{p}_{br}}{2}.
\end{equation}

From this point onward, $(p_x,p_y)$ denotes the image coordinates associated with the detected target.

Once the target has been identified, the module combines the selected image coordinates with the corresponding depth information provided by the RGB-D sensor to estimate the target position in three-dimensional space. Since depth measurements may be noisy or affected by missing values, the depth assigned to the target is computed as the minimum valid depth value within a square neighborhood of size $(2r+1)\times(2r+1)$ centered at $(p_x,p_y)$. Invalid measurements, such as zero or infinite depth values, are discarded during this process to improve the robustness of the estimation,
\begin{equation}
d=
\min_{\substack{
u \in [p_x-r,p_x+r]\\
v \in [p_y-r,p_y+r]\\
0 < D(u,v) < \infty
}}
D(u,v),
\end{equation}

\noindent where $D(u,v)$ denotes the depth image and $r$ defines the size of the neighborhood considered around the center pixel. Selecting the minimum valid depth value increases the probability of associating the measurement with the detected object rather than with background elements.

Since the objective is to approach the target rather than reach its geometric center, an offset $\delta$ can be applied to the estimated depth. The corrected depth is computed as $d' = d - \delta$, where $d$ is the measured depth and $d'$ is the adjusted depth used for goal generation. The value of $\delta$ depends on the target object; for example, a larger structure such as a bus stop may require a greater offset than a person or a mailbox. This adjustment allows the robot to stop at a suitable distance from the target instead of navigating towards its center.

Given the target image point $\mathbf{p}=(p_x,p_y)$ and its associated adjusted depth $d'$, the corresponding 3D position in the camera reference frame can be recovered using the pinhole camera model \cite{hartley2003multiple}:
\begin{equation}
X = \frac{(p_x-c_x)d'}{f_x}, \quad
Y = \frac{(p_y-c_y)d'}{f_y}, \quad
Z = d'.
\end{equation}
where $(c_x,c_y)$ represent the coordinates of the principal point of the camera, corresponding to the optical center of the image, and $(f_x,f_y)$ are the focal lengths expressed in pixels, obtained from the intrinsic calibration parameters of the camera.
The resulting point
\begin{equation}
\mathbf{P}_c=(X,Y,Z)\in\mathbb{R}^{3}
\end{equation}
defines the local position of the detected target with respect to the camera reference frame. Using the robot localization information together with the coordinate transformations available through the ROS~2 TF framework, this point is transformed into the global map frame according to
\begin{equation}
\mathbf{P}_m =
{}^{m}\mathbf{T}_{c}\mathbf{P}_c,
\end{equation}
where ${}^{m}\mathbf{T}_{c}\in SE(3)$ denotes the rigid-body transformation between the camera and map reference frames obtained from the ROS~2 TF tree. The resulting point $\mathbf{P}_m\in\mathbb{R}^{3}$ defines the target position in the global map reference frame and constitutes the navigation goal subsequently forwarded to the Movement Module for execution.

\subsection{Movement Module}

The Movement Module acts as the interface between the
high-level navigation goals generated by the Semantic Perception Module and the robot navigation stack. Its primary
responsibility is to transform the navigation objective received
from the Communication Module into an executable navigation action and monitor its execution.

Upon receiving a goal position expressed in the global map frame, the Movement Module sends the corresponding navigation request to the ROS~2 Navigation Stack (Nav2) \cite{macenski2020marathon}. Nav2 computes a feasible path to the target location and generates the control commands required to execute the navigation task.

The proposed framework assumes that a functional low-level controller is available on the robotic platform. Consequently, the Movement Module operates independently of the specific locomotion system of the robot.

Path planning and obstacle avoidance are delegated to Nav2. A global planner computes a collision-free path using the global costmap, while a local planner continuously adapts the robot motion using the local costmap to account for static and dynamic obstacles. These costmaps are generated from the robot's onboard sensors, such as LiDAR and depth cameras. By leveraging these capabilities, the Movement Module focuses on connecting the semantic navigation objective generated by the proposed system with the robot navigation infrastructure, enabling the robot to safely reach the goal position estimated by the Semantic Perception Module.

\begin{figure*}[t]
    \centering
    \begin{tabular}{ccc}
        \includegraphics[width=0.30\textwidth,height=3.1cm]{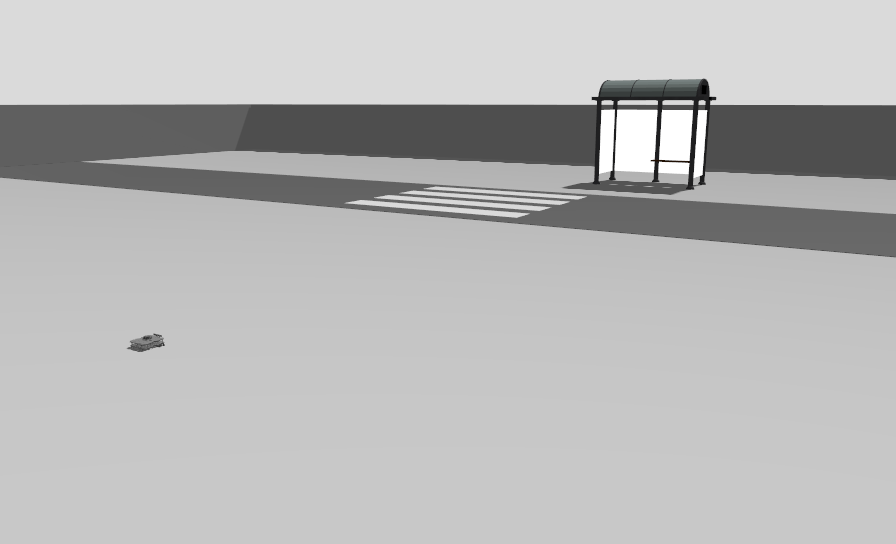} &
        \includegraphics[width=0.30\textwidth,height=3.1cm]{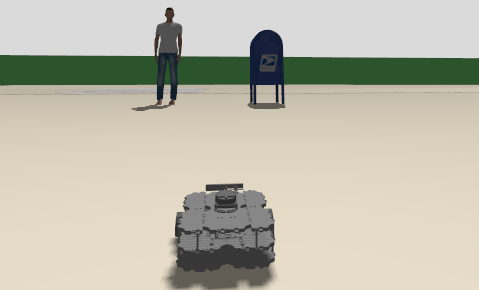} &
        \includegraphics[width=0.30\textwidth,height=3.1cm]{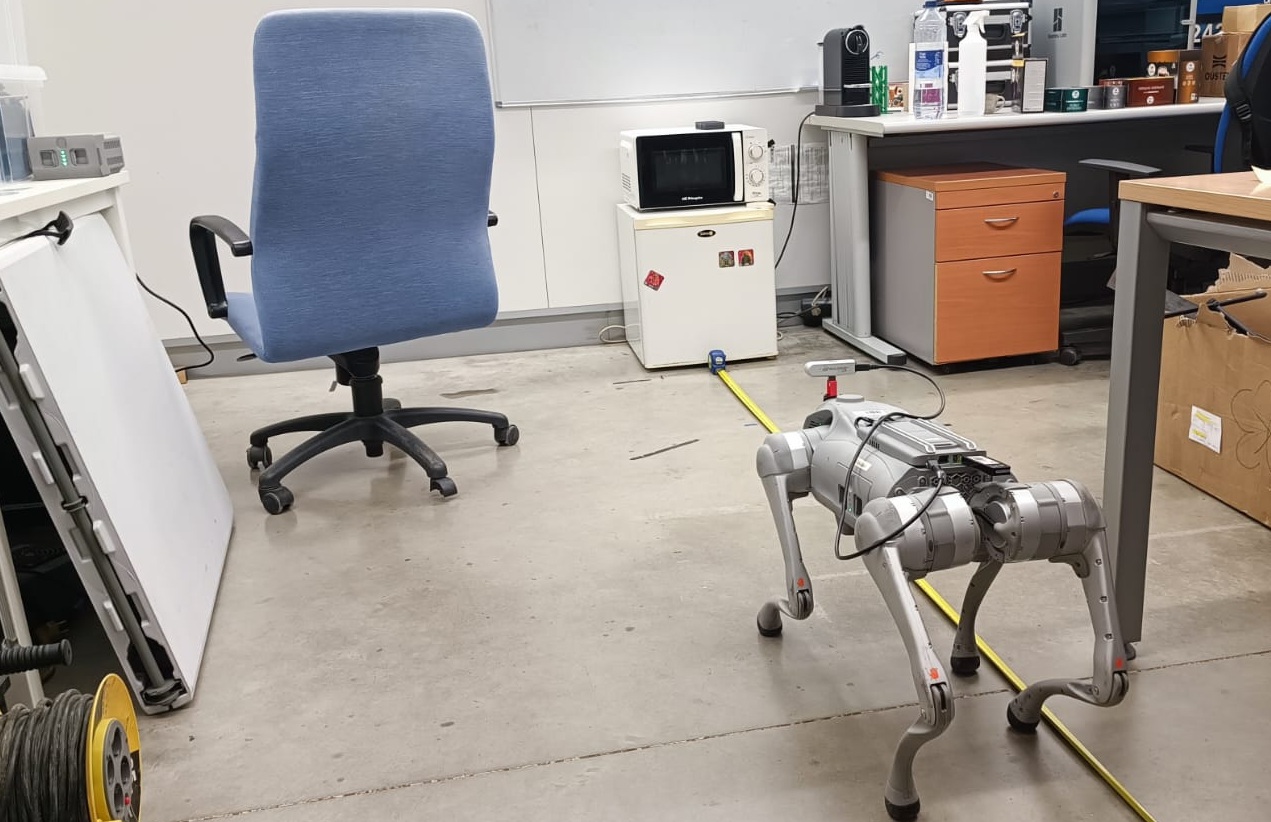} \\
    \end{tabular}

    \caption{Overview of the experimental scenarios. (Left) Experiment~1 evaluates the Semantic Perception Module by estimating navigation goals for a bus stop target in simulation. (Center) Experiment~2 evaluates the complete language-driven navigation framework, where the robot must distinguish between different target categories (person and mailbox), identify the intended target, and autonomously navigate towards it. (Right) Experiment~3 evaluates the Semantic Perception Module on a real robotic platform using a Unitree Go2 equipped with an Intel RealSense RGB-D camera, validating the transferability of the proposed approach to real-world environments.}
    \label{fig:experiments_overview}
\end{figure*}

\section{Experiments}
\label{sec:experiments}
The proposed framework was evaluated through three complementary experiments. The first experiment assesses the accuracy of the Semantic Perception Module in simulation by comparing the generated navigation goals against ground-truth target locations. The second experiment evaluates the performance of the complete language-driven navigation pipeline, including semantic perception, goal generation, and autonomous navigation. Finally, the third experiment presents a real-world proof of concept to validate the Semantic Perception Module on a real robotic platform, demonstrating the transferability of the proposed approach to real-world environments.

The simulation experiments were conducted in Gazebo using a TurtleBot3 Waffle integrated with ROS~2 and Nav2. The real-world experiment was performed using a Unitree Go2 equipped with an Intel RealSense RGB-D camera.

\subsection{Experiment 1: Semantic Perception Evaluation}

The objective of this experiment is to evaluate the ability of
the Semantic Perception Module to transform natural language
instructions into accurate navigation goals. To isolate the
performance of the perception pipeline, a single target object,
a bus stop, is considered throughout the experiment. During
this evaluation, no navigation action is executed, allowing the
perception stage to be assessed independently of the robot
navigation stack

To evaluate the robustness of the proposed approach, four
trials were conducted using different natural language formulations referring to the same target object and different initial
robot poses were considered, resulting in varying viewing
angles and distances to the target. This configuration allows the
robustness of the Semantic Perception Module to be evaluated
against both linguistic variations and changes in the robot
viewpoint while maintaining the same semantic target.
The evaluated prompts were:

\begin{enumerate}
    \item \textit{``Can you go to the bus stop, please?''}
    \item \textit{``Find a bus stop''}
    \item \textit{``Reach the bus stop in front of you''}
    \item \textit{``Would you be able to reach the bus stop?''}
\end{enumerate}

For each trial, the system receives a natural language command and estimates a navigation goal using the procedure described in Section~\ref{sec:semantic}. The generated goal incorporates the depth offset parameter $\delta$, which is used to position the robot at an appropriate distance from the target. For the bus stop, the offset was set to $\delta = 0.6$~m. The estimated goal position is then compared with a ground-truth navigation goal.

The localization error is defined as the Euclidean distance between the ground-truth navigation goal in the map frame, $(x_{gt},y_{gt})$, and the goal estimated by the Semantic Perception Module, $(x_{est},y_{est})$:

\begin{equation}
\label{e1_error}
e_{goal}
=
\sqrt{
(x_{gt}-x_{est})^{2}
+
(y_{gt}-y_{est})^{2}
}.
\end{equation}
Table~\ref{tab:bus_stop} summarizes the results obtained for the different prompts and initial robot poses.
\begin{table}[t]
\centering
\caption{Semantic perception results for the bus stop (GT: 18, 0)}
\label{tab:bus_stop}
\begin{tabular}{cccc}
\toprule
Test & Initial Pose [m] & Estimated Goal [m] & Error [m] \\
\midrule
1 & (0.06, -0.06)  & (18.68, 0.03) & 0.69 \\
2 & (2.06, 0.94)   & (18.68, 0.05) & 0.69 \\
3 & (-2.94, -0.06) & (18.68, 0.01) & 0.68 \\
4 & (1.06, -2.06)  & (18.68, 0.08) & 0.69 \\
\midrule
Mean & -- & -- & 0.68 \\
\bottomrule
\end{tabular}
\end{table}

\begin{figure}[!b]
    \centering
    \includegraphics[width=0.40\columnwidth]{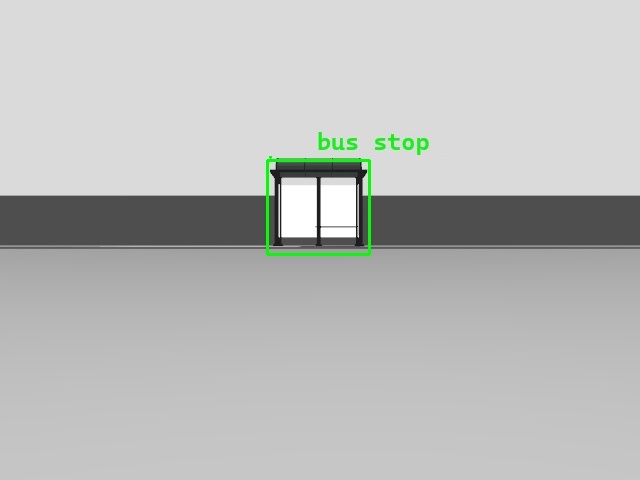}
    \includegraphics[width=0.40\columnwidth]{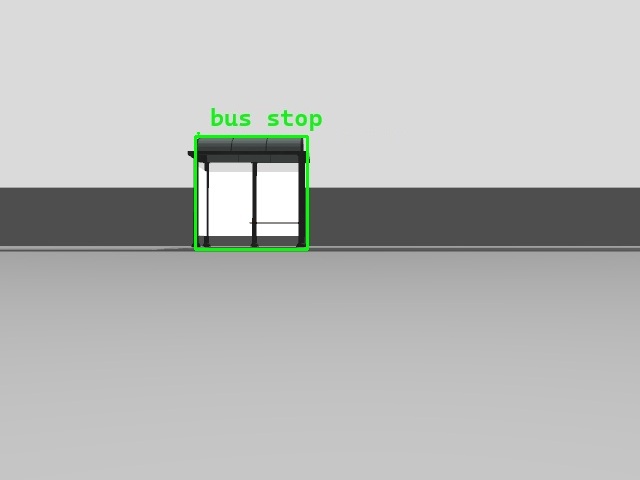}
    \caption{Representative RGB images from different experimental trials. Different initial robot poses produce different viewpoints of the same bus stop target. Despite variations in perspective, distance, and natural language formulation, the VLM consistently identifies the target object.}
    \label{fig:bus_stop_detections}
\end{figure}

Figure~\ref{fig:bus_stop_detections} presents qualitative results corresponding to the experiments reported in Table~\ref{tab:bus_stop}. The robot starts from different initial poses, resulting in varying viewpoints and scene compositions. Despite these changes, the system  identifies the target and generates a bounding box around it. The center of the bounding box is then used by the Semantic Perception Module to estimate the target's three-dimensional position and generate the corresponding navigation goal.

The results show that the proposed Semantic Perception Module can reliably identify the target despite variations in viewpoint and language formulation. An average localization error of $0.68$~m was obtained across all experiments. It should be noted that this error is measured with respect to the ground-truth target position, whereas the generated goal is intentionally shifted using an offset of $\delta=0.6$~m to place the robot at a practical navigation distance from the bus stop.

\begin{figure}[b]
    \centering
    \begin{tabular}{cc}
        \includegraphics[width=0.48\columnwidth]{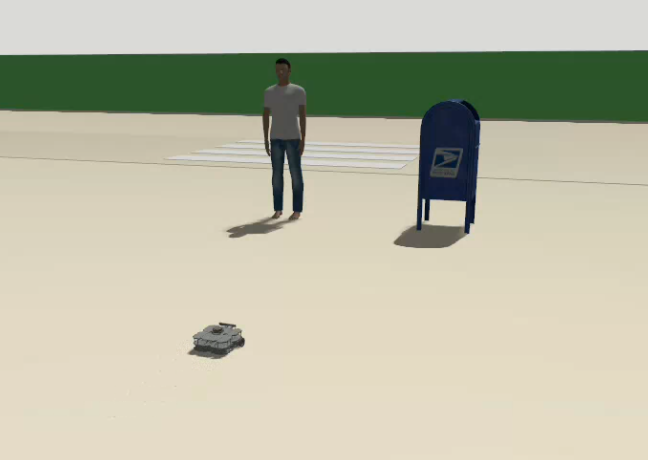} &
        \includegraphics[width=0.48\columnwidth]{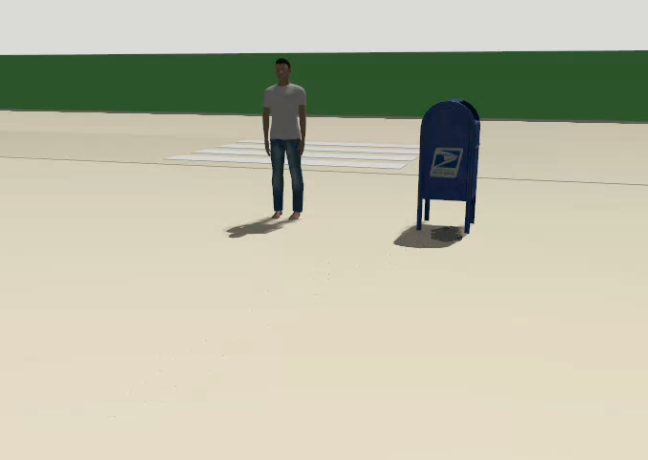} \\
    \end{tabular}
    \caption{Representative navigation sequence obtained during one of the end-to-end experiments. User requested: ``I need to send a letter. Can you give me a way to send it?''. From left to right, the robot progresses from its initial position towards the target. The sequence demonstrates the successful execution of the generated navigation goal and the ability of the framework to transform natural language instructions into autonomous navigation behaviors.}
    \label{fig:exp2_navigation_sequence}
\end{figure}

\subsection{Experiment 2: End-to-End Navigation Evaluation}

The objective of this experiment is to evaluate the complete language-driven navigation framework. In contrast to Experiment~1, the robot executes the navigation goals generated by the Semantic Perception Module using the Movement Module and the Nav2 navigation stack.

To assess the robustness of the proposed approach, six experiments were conducted considering two semantic targets (aperson and a mailbox), different natural language formulations,
and different initial robot poses. For each trial, the system
receives a natural language command, identifies the referenced
target, generates a navigation goal, and autonomously navigates towards it.
For each trial, the system receives a natural language
command, generates a navigation goal using the Semantic
Perception Module, and executes the navigation task through
Nav2. Once the navigation process is completed, the final robot
pose is compared with the ground-truth target position.
Let I = (xi
, yi) denote the initial robot pose, R = (xr, yr)
the final robot pose after navigation, and G = (xgt, ygt) the
ground-truth position of the target object. The navigation error,
enav, is defined as the Euclidean distance between R and G:

\begin{table}[t]
\centering
\caption{End-to-end navigation evaluation prompts and target identification.}
\label{tab:semantic_prompts}
\footnotesize
\renewcommand{\arraystretch}{1.15}
\setlength{\tabcolsep}{4pt}
\begin{tabular}{c p{3.2cm} p{3.2cm} c}
\toprule
ID & Prompt & Response & Target \\
\midrule
1 & ``Reach the person'' & ``I have located the person.'' & Person \\
2 & ``Find a male and reach him'' & ``I found a male on the left side of the image.'' & Person \\
3 & ``My father is next to a mailbox. Can you take me to him?'' & ``I found your father next to the mailbox.'' & Person \\
4 & ``Reach the mailbox'' & ``I have located the mailbox.'' & Mailbox \\
5 & ``I need to send a letter. Can you give me a way to send it?'' & ``You can send the letter by placing it in the mailbox.'' & Mailbox \\
6 & ``You have to find a mailbox'' & ``I found a mailbox.'' & Mailbox \\
\bottomrule
\end{tabular}
\end{table}

Let $I=(x_i,y_i)$ denote the initial robot pose, $R=(x_r,y_r)$ the final robot pose after navigation, and $G=(x_{gt},y_{gt})$ the ground-truth position of the target object. The navigation error, $e_{nav}$, is defined as the Euclidean distance between $R$ and $G$:

\begin{equation}
\label{e2_error}
e_{nav}
=
\sqrt{
(x_{gt}-x_r)^2 +
(y_{gt}-y_r)^2
}.
\end{equation}

In addition to the navigation error, the execution time and traveled distance were recorded. Furthermore, the natural-language response generated by the VLM was included in the evaluation, as it provides feedback in natural language to the user regarding the identified target and allows the interpreted destination to be verified before navigation begins.

\begin{table}[t]
\centering
\caption{Quantitative navigation metrics for the evaluation scenarios. $I=(x_i,y_i)$ denotes initial robot pose, $R=(x_r,y_r)$ final robot pose, and $G=(x_{gt},y_{gt})$ ground-truth position.}
\label{tab:semantic_metrics}
\footnotesize
\renewcommand{\arraystretch}{1.15}
\setlength{\tabcolsep}{3pt}
\resizebox{\columnwidth}{!}{%
\begin{tabular}{c c c c c c c}
\toprule
ID & $I$ [m] & $R$ [m] & $G$ [m] & $e_{nav}$ [m] & Time [s] & Distance [m] \\
\midrule
1 & (0.07,-0.05) & (4.44,0.87) & (5.00,1.00) & 0.575 & 21 & 4.86 \\
2 & (-1.93,-2.05) & (4.50,0.80) & (5.00,1.00) & 0.539 & 34 & 8.24 \\
3 & (-0.93,0.95) & (4.36,0.88) & (5.00,1.00) & 0.651 & 27 & 5.96 \\
4 & (0.07,-0.05) & (4.31,-0.75) & (5.00,-1.00) & 0.734 & 24 & 4.96 \\
5 & (-1.93,1.95) & (4.21,-0.83) & (5.00,-1.00) & 0.808 & 30 & 7.81 \\
6 & (-0.93,-1.05) & (4.09,-0.92) & (5.00,-1.00) & 0.914 & 20 & 5.47 \\
\midrule
\multicolumn{4}{c}{Mean} & 0.704 & 26.0 & 6.22 \\
\bottomrule
\end{tabular}%
}
\end{table}

Table~\ref{tab:semantic_metrics} summarizes the obtained results. Besides the navigation metrics, the Table~\ref{tab:semantic_prompts} reports the original user prompt, the response generated by the VLM, the target object. For this experiment, the depth offset parameter was set to $\delta=0.3$~m. The navigation error $e_{nav}$ is computed according to Eq.~(\ref{e2_error}), allowing a direct comparison between the desired destination and the position ultimately reached by the robot.

It should be noted that the objective is not to reach the exact target coordinates, which may correspond to the geometric center of the object, but rather to safely approach the desired destination. Consequently, non-zero values of $e_{nav}$ are expected due to the applied depth offset, the dimensions of the target object, and the obstacle avoidance behavior of Nav2. This effect is particularly noticeable for the mailbox target, which errors are larger than the person target due to its larger size.

In addition to the quantitative evaluation, qualitative results
were collected to illustrate the behavior of the proposed framework during navigation. Figure 2 in Sec.III-C shows the firstperson view acquired by the robot during target identification,
while Fig. 5 presents a third-person view of a representative
navigation trial. Together, these figures illustrate the complete
pipeline, from semantic target detection to the execution of
the generated navigation goal.

The results demonstrate that the proposed framework successfully combines semantic perception and autonomous navigation. Across all experiments, the robot correctly identified
the intended target, generated meaningful natural-language
feedback, and reached the desired destination with an average
navigation error of 0.70 m. Furthermore, the robot traveled an
average distance of 6.22 m and required an average execution
time of 26 s, confirming the feasibility of the proposed
language-driven navigation framework.

\subsection{Experiment 3: Real-World Validation}

The objective of this experiment is to provide a proof of concept for the Semantic Perception Module on a real robotic platform. The evaluation was conducted using a Unitree Go2 equipped with an Intel RealSense RGB-D camera.

\begin{figure}[ht]
    \centering
    \includegraphics[width=0.48\linewidth]{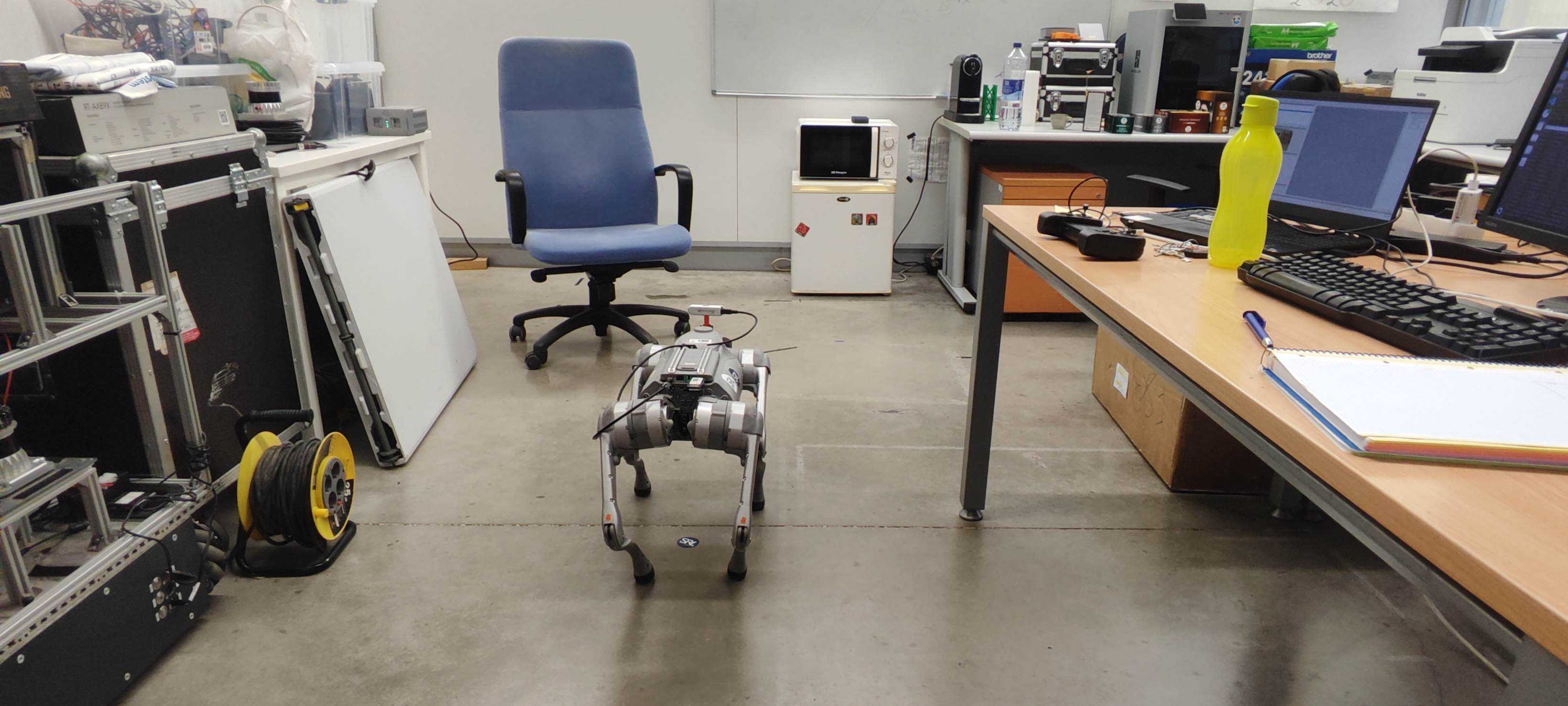}
    \hfill
    \includegraphics[width=0.48\linewidth]{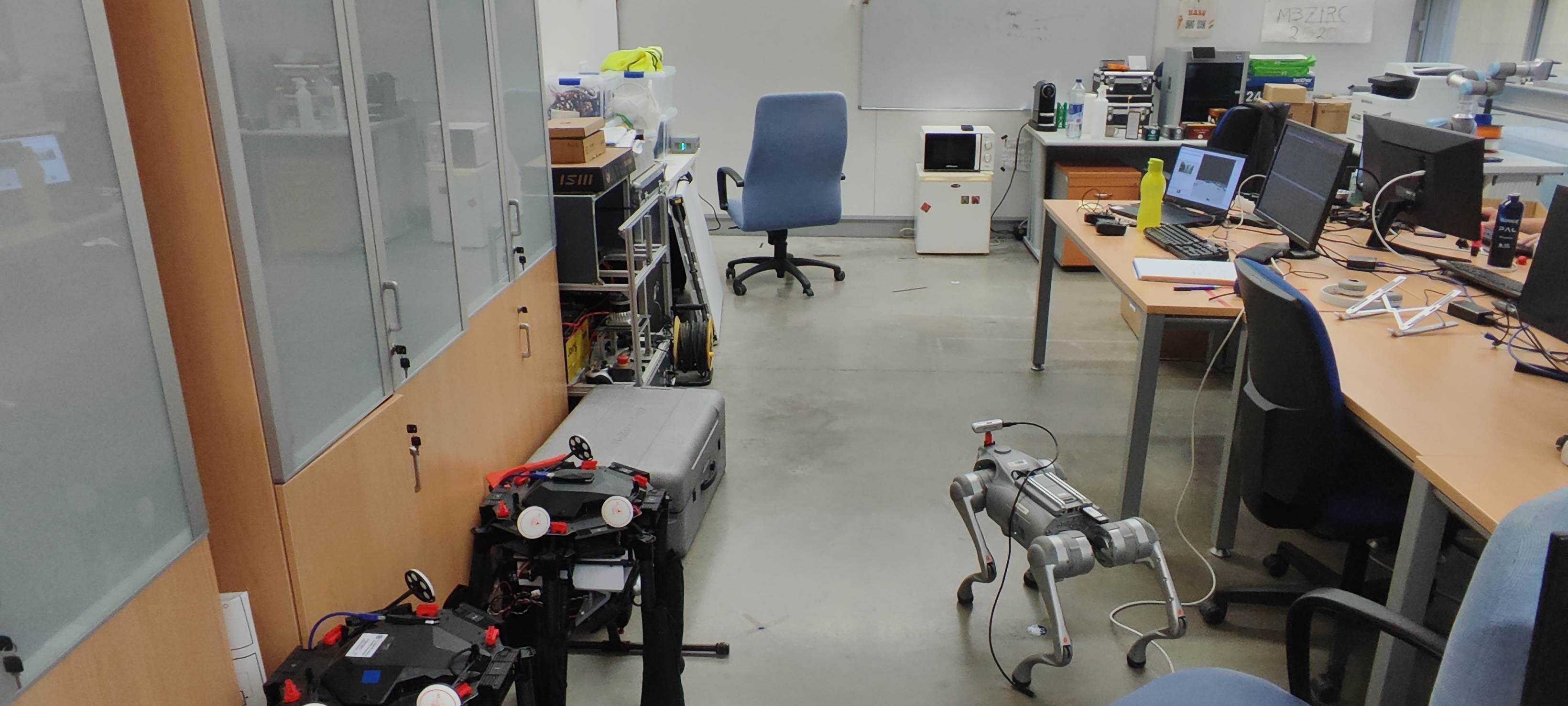}
    \caption{Real-world experimental setups used in Experiment~3. The Unitree Go2 equipped with an Intel RealSense RGB-D camera is shown in two different configurations. Different robot poses, target objects, and observation angles were considered to evaluate the robustness of the Semantic Perception Module under real-world conditions.}
    \label{fig:exp3_layouts}
\end{figure}

Figure~\ref{fig:exp3_layouts} presents the two experimental layouts used during the evaluation. The considered targets were a microwave and a chair, while both direct and contextual natural language prompts were employed.

The evaluated prompts were:

\begin{enumerate}
    \item \textit{``Reach the microwave''}
    \item \textit{``I need something to heat up my food''}
    \item \textit{``Reach the chair''}
    \item \textit{``I want to sit''}
\end{enumerate}

For each trial, the robot receives a natural language request, identifies the referenced object using the Semantic Perception Module, and estimates its position from RGB-D data. Ground-truth target positions were obtained through direct measurements in the environment and used to compute the localization error according to Eq.~(\ref{e1_error}). 
\begin{figure}[ht]
    \centering
    \begin{tabular}{cc}
        \includegraphics[width=0.40\linewidth]{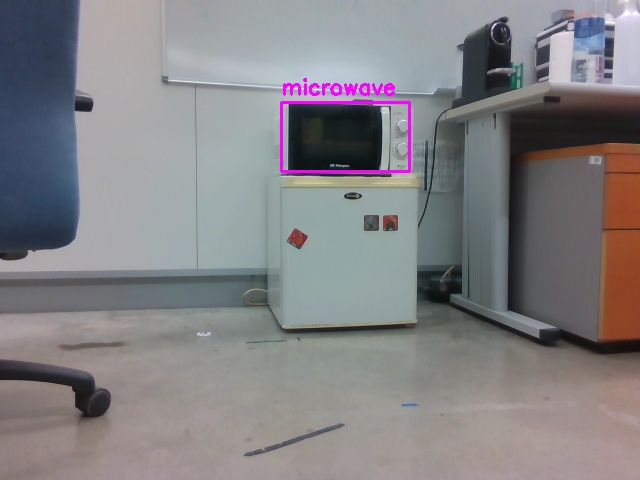} &
        \includegraphics[width=0.4\linewidth]{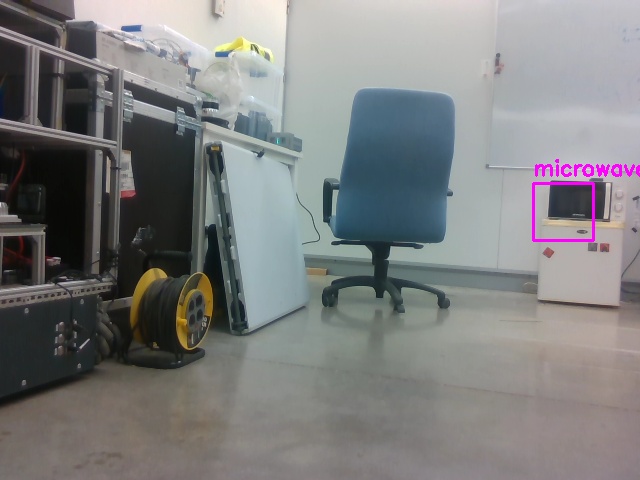} \\
        \includegraphics[width=0.4\linewidth]{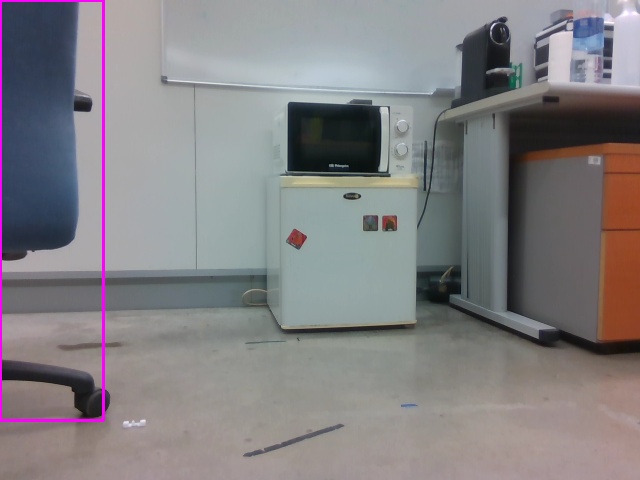} &
        \includegraphics[width=0.4\linewidth]{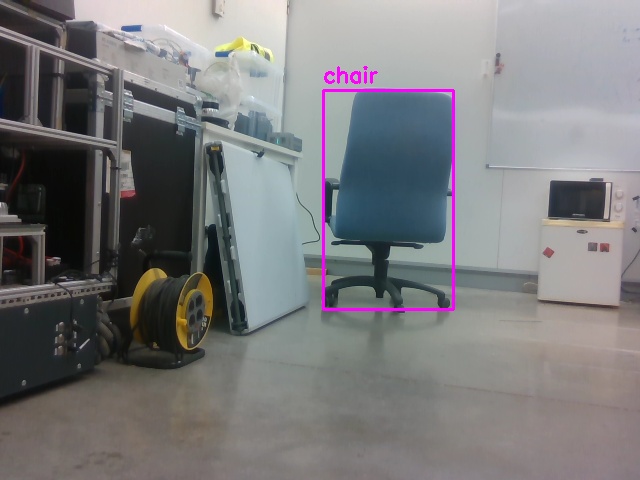} \\
    \end{tabular}
    \caption{Representative first-person views obtained during the real-world evaluation. The images show the bounding boxes generated by the VLM for two targets: microwave (top row) and chair (bottom row) observed from different viewpoints. These detections are subsequently used by the Semantic Perception Module to estimate the target position from RGB-D data.}
    \label{fig:real_world_results}
\end{figure}

Table~\ref{tab:real_world_results} summarizes the obtained results. Tests 1--4 correspond to prompts 1--4 listed above, respectively.

\begin{table}[b]
\centering
\caption{Real-world semantic perception evaluation results. Coordinates are expressed in the robot local reference frame.}
\label{tab:real_world_results}
\small
\begin{tabular}{ccccc}
\toprule
Test & Target & Estimated [m] & Ground Truth [m] & Error [m] \\
\midrule
1 & Micro & $(2.12,0.34)$ & $(2.25,0.33)$ & 0.13 \\
3 & Micro & $(4.20,0.00)$ & $(3.92,0.17)$ & 0.33 \\
2 & Chair      & $(1.58,-0.44)$ & $(0.99,-0.04)$ & 0.71 \\
4 & Chair      & $(3.45,-0.94)$ & $(3.01,-0.19)$ & 0.87 \\
\midrule
Mean & -- & -- & -- & 0.51 \\
\bottomrule
\end{tabular}
\end{table}

Figure~\ref{fig:real_world_results} presents representative qualitative results obtained with the Unitree Go2 platform. Four first-person observations are shown, corresponding to the two target categories viewed from different robot poses and scene configurations.  Despite variations in viewpoint, distance, and scene composition, the
VLM consistently identifies the requested target and generates
accurate bounding boxes, which are subsequently used by the
Semantic Perception Module for geometric localization.

The results demonstrate that the proposed Semantic Perception Module generalizes successfully to real-world environments. Across all evaluated scenarios, the system correctly identified the requested objects and obtained an average localization error of $0.51$~m. Since this value is comparable to the applied offset ($\delta=0.5$~m), most of the measured error can be attributed to the intentional displacement used to position the robot at a practical distance from the target.
\section{Conclusions}
\label{sec:conclusions}

This paper presented a language-driven navigation framework that combines Vision-Language Models (VLMs), RGB-D perception, and autonomous navigation to enable mobile robots to interpret natural language requests and navigate towards the referenced targets. By integrating semantic understanding with geometric localization, the proposed system transforms high-level instructions into executable navigation goals without requiring explicit coordinates, enabling its use to non-expert users.

The framework was validated through three complementary experiments, including two simulation-based evaluations and real-world proof of concept for semantic perception. Across all experiments, the system successfully identified the intended targets, generated meaningful natural-language feedback, and produced reliable navigation goals.

Overall, the results demonstrate the potential of VLMs to bridge natural language interaction, semantic perception, and autonomous navigation. Future work will focus on 
make the framework work fully in real-world environments, extending the framework to support more complex multi-step tasks, improving goal generation through adaptive offset estimation.


\bibliographystyle{IEEEtran}
\bibliography{IEEEabrv,main}

\end{document}